\documentclass[11pt]{article}

\usepackage{amssymb,latexsym,amsmath,amsbsy}
\usepackage{graphicx}
\usepackage{cite}
\usepackage{hyperref}
\usepackage{scrextend}

\newtheorem{theorem}{Theorem}

\newcommand{\minuscule}{\@setfontsize\minuscule{4}{5}}

\newcommand{\BE}{\begin{equation}}
\newcommand{\EE}{\end{equation}}
\newcommand{\BQ}{\begin{equation} \begin{array}{c}}
\newcommand{\EQ}{\end{array}\end{equation}}
\newcommand{\BT}{\begin{theorem}}
\newcommand{\ET}{\end{theorem}}

\newcommand{\BL}{\begin{lstlisting}}
\newcommand{\EL}{\end{lstlisting}}

\newcommand{\ZB}{\mathbb{Z}}

\usepackage{listings}
\usepackage{color}

\begin{document}

\begin{center}
\noindent
{\Large \bf
  $XOR_p$\\
  A maximally intertwined p-classes problem\\
  used as a benchmark with built-in truth for\\
  neural networks gradient descent optimization.\\ 
}

\vskip 10mm

{\bf  Danielle Thierry-Mieg, Jean Thierry-Mieg}\\[2mm]
National Center for Biotechnology Information,\\
National Library of Medicine, National Institutes of Health,\\
8600 Rockville Pike, Bethesda MD20894, USA.\\
 E-mail: mieg@ncbi.nlm.nih.gov \\[2mm]
\end{center}

\vskip 10mm





\begin{abstract}
A natural p-classes generalization of the eXclusive OR problem,
the subtraction modulo p, where p is prime, 
is presented and solved using a single fully connected hidden layer 
with p-neurons. Although the problem is very simple, the landscape is
intricate and challenging and represents an interesting benchmark
for gradient descent optimization algorithms. Testing 9
optimizers and 9 activation functions up to $p$ = 191, the method 
converging most often and the fastest to a perfect classification
is the Adam optimizer combined with the ELU activation function.
\end{abstract}



\section{Introduction}

As discussed in 1969 by Marvin Minsky and Seymour Papert \cite{[1]}
the $XOR$ (exclusive-or) problem is a seminal example, arguably the simplest operation
that cannot be solved by a single layer neural network \cite{[2]}.
As found later, the solution requires a second hidden layer and
a non linear activation function. Yet, 
the number of iterations needed to train the network using
the simple gradient descent algorithm is surprisingly high
(see table 1), especially if we remember that
we are just trying to see if two bits, $a$ and $b$, are unequal:
indeed $a\;XOR\;b$ is TRUE if and only if $a \neq b$.

The difficulty comes from the fact that the classes are intertwined.
As shown in figure 1, the true and false classes cannot 
be separated by a linear equation.
Indeed, we are not trying to sort the pairs $(a,b)$ by looking separately at 
properties of the objects $a$ or $b$, but rather we sort the four $(a,b)$
configurations based on the relative values of $a$ and $b$. In other words,
the problem is indecomposable in the sense that no sub-classification of the
values of $a$ helps solving the complete problem.

We propose to define a classification problem
which generalizes, for an arbitrarily large number of classes, 
the fundamental indecomposable nature of the $XOR$ problem.
On this theoretical framework, we then evaluate
how often and how fast the classical gradient descent optimizers
converge to the known absolute minima, and the interplay
with the choice of the activation function.

\begin{center}
\begin{figure}[h!]
  \includegraphics[width=15cm,height=5cm]{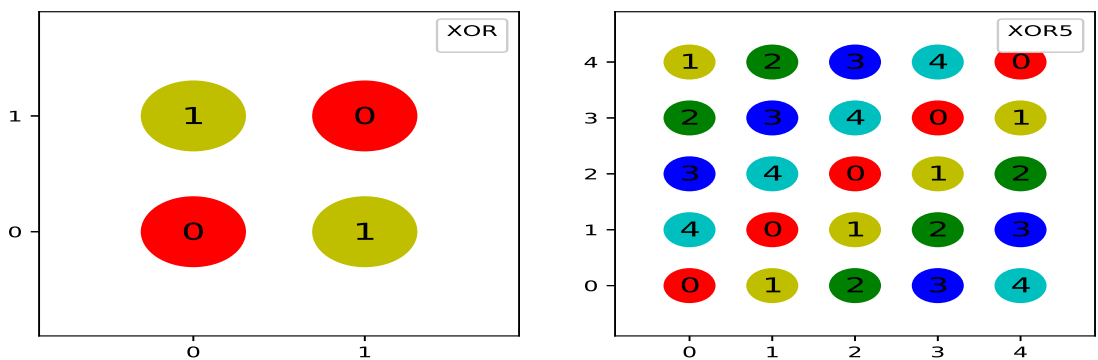}
  \label{fig:xor}
  \caption{Partition of the square in $p$ classes. a:$XOR$, b:$XOR_5$.
No single line can separate the classes in 2 subsets.
 $2 p - 2$ lines are needed to separate all the classes.
}
\end{figure}
\end{center}

\section{Generalization of the $XOR$ problem to p classes}

We wish to define a maximally intertwined
p-classes classification problem, where no pair of classes can be
separated by a linear equation and  for which any number of training and test sets can be
synthesized easily.
A good candidate can be constructed from
classical considerations in group theory.
Consider the exact sequence:
\BE
0 \longmapsto \ZB_p \longmapsto \ZB_p \times \ZB_p \longmapsto \ZB_p \longmapsto 0
\EE
where $\ZB_p$ denotes the group of integers modulo $p$,
each arrow represents a linear homomorphism preserving the addition modulo $p$,
and each double map yields a zero. It may be implemented, by mapping the left $0$ to the
$0$ of the first  $\ZB_p$, then mapping any element $a$ of the left $\ZB_p$ 
to the diagonal element $(a,a)$ of  $\ZB_p \times \ZB_p$. 
Then mapping any element $(a,b)$ of  $\ZB_p \times \ZB_p$ to
$c = a - b \;\;modulo\;p$, and finally mapping the last  $\ZB_p$
to $0$.  These maps commute with the rules of the addition modulo $p$
and one can verify that the central double map yields zero $a \longmapsto (a,a) \longmapsto 0$.
When $p$ is prime, the group $\ZB_p$ is simple and the exact sequence
is indecomposable. 

If $p = 2$, we can verify that $(0,0)$ and $(1,1)$ map to $0$, 
and that $(1,0)$ and $(0,1)$ map to $1$,
therefore the construction implements $XOR$ and the
subtraction modulo $p$ can be seen as a natural p-classes generalization
of the eXclusive OR. 

Please notice that when $p = 2$, $a - b$ modulo $2$ is identical to $a + b$.
More generaly, for any $p$ one can replace $a - b$ modulo $p$ by $a + b'$ modulo $p$
where $b' = p - b$. This modification amounts to swapping the orientation of the
enumeration of the second set: i.e. in the case $p=5$, $b$ is in the
set $(0,1,2,3,4)$ and $b'$ in $(0,4,3,2,1)$. For this reason, although $a - b$
is antisymetric in $(a,b)$, the neural networks constructed below
to solve ($a\;\ZB_p\;b$) and ($b\;\ZB_p\;a$) are equivalent.

Figure 1 present the different classes in the
discrete cases $XOR$ and $XOR_5$ and helps to see the intertwined nature of
these partitions of the square. One can verify the $2p - 2$ oblique lines
are needed to separate all the classes. Despite the apparent regularity of the 
figure, this number is maximal since it is equal to the number of lines
needed to create a horizontal-vertical grid separating all the cells.
One may also verify that no single line separates
any pair of classes, say the (2) and the (3), or any 2 complementary sets
of classes, say the (0,1,4) versus the (2,3). One may also define a continuous 
pattern of oblique classes, say $C_5$,
where $C_p$ is defined by the equation $c = int(p * (a - b) + .5)$ modulo $p$,
where $a$ and $b$ are real numbers in the interval $[0,1]$ and
$c$ is in  $\ZB_p$. 

To summarize, we propose to consider the partition
of pairs of integers $(a,b)$ modulo $p$ into $p$ classes according to 
their difference $c = a - b$ modulo $p$, 
as the archetype of a complex decision problem.

\section {A $\ZB_p$ neural network with a single hidden layer}

Since the $XOR$ problem can be solved with a single hidden layer, we
propose to encode $a$, $b$ and $c$ as p-dimensional 1-hot vectors,
e.g. the numbers 3 and 4 of $\ZB_5$ are represented respectively
by the vectors $(0,0,0,1, 0)$ and  $(0,0,0,0,1)$,  
and to construct a single hidden layer with dimension $(2p,p)$ followed by
a fully connected  $(p,p)$ layer with softmax and cross entropy cost function \cite{[3]}.

This network is highly symmetric. Even if we fix the representations of
$a$, $b$ and $c$ in the input and output layers, any pair of neurons in the hidden
layer may be swapped, yielding $p!$ equivalent configurations.
In addition, there are many possible classes of solutions. 
Recall that the $(p=2)$ classic eXclusive OR can be written
in several different ways, for example as  
$(a \wedge \neg b) \vee (\neg a \wedge b)$ 
but also as $(a \vee b) \wedge \neg (a \wedge b)$.
For higher $p$, the number of ways to compute $ a - b \;\;modulo\;p$
as a 2-step process increases rapidly
so the number of distinct networks solving the $\ZB_p$ problem
becomes very large for large $p$.

Intuitively, the landscape resembles at the same time
a hall of mirrors, each mirror swapping a pair of neurons,
and one of these non-trivial wood puzzles where
all pieces may have slightly different shapes, representing our $p$
slightly different neurons, and yet globally fill a simple pleasing
pattern, the subtraction modulo $p$.

\section { Gradient descent optimization}

Although the mathematical problem is very simple, its implementation
becomes tricky just because we follow the standard paradigm
of neural networks. We are trying to use a real matrix of
dimensions $(2p,p)$ and $p$ biases in the hidden layer, plus a real matrix
of dimension $(p,p)$ and $p$ biases in the output layer. In total, we would like
to automatically adjust a set of $3 p^2 + 2 p$ real parameters
by the method of gradient descent  \cite{[4]}, and require a perfect result
for $p^2$ configurations of the input variables $(a,b)$.
The game is to find the gradient descent optimization algorithm
that will converge on average most rapidly to a correct solution.

There are many possibilities. We systematically tried seven
algorithms offered in TensorFlow \cite{[5]},
namely the simple gradient descent \cite{[6]}, often nicknamed Vanilla, the
momentum method \cite{[7]} with or without Nesterov correction \cite{[8]},
the Adagrad optimizer \cite{[9]} and its gemeralizations Adadelta \cite{[10]},
RMSProp \cite{[11]} and Adam \cite{[12]}. We then added two
recent optimizers L4Adam and L4Mom \cite{[13]} which leverage
the Adam and Momentum optimizer with a 'Local Linear optimaL Learning-rate'
policy.
As recalled in the introduction, the network cannot work
unless the hidden layer contains a non linear activation function \cite{[2]}.
There are again many choices and we tested 9 methods (figure 2).
Three functions are differentiable,  the classic sigmoid function,
the hyperbolic tangent (TANH) and the intermediate exponential-linear (ELU) 
function \cite{[14]}. Following the instructions in \cite{[4]} page 283,
we then tested 6 piecewise linear functions which are in principle
faster to compute.
We first considered the linear rectifier (RELU)  \cite{[15]},
the leaky RELU \cite{[16]} with slopes 0.2, and the bounded RELU  \cite{[17]} saturating at 2, 
i.e. a piece-wise linear approximation of the sigmoid.
We then added two piece-wise linear approximations
of ELU, either using 2 lines: LELU defined as $y = max(-1,x)$,
or three: L3ELU, defined by 3 optimal linear segments $y = max(-1,0.231 x - 0.387,x)$;
where the equation of the central segment was chosen to minimize
the square distance $\int (L3ELU(x) - ELU(x))^2\;dx$.
Finally, we added a leaky linearized ELU: LLELU defined as
$y = max(x, -1 + (x+1)/5)$.
Notice that a fast
generic way to compute a piece-wise linear approximation
of any convex function is simply to compute as in the examples above 
the max of the linear functions defining the different segments.

These functions, presented in figure 2,  were chosen to assess if the difference
between ELU and RELU is caused by replacing the corner of the RELU by the differentiable
bend of the ELU, or rather if the difference comes from the left asymptote
being at $-1$ in ELU and at $0$ in RELU, thus allowing to differentiate
small and large negative values. L3ELU is similar to the
recently defined GRELU activation function \cite{[18]}, both
converge to a negative constant for large negative $x$ values, 
and not to zero as in RELU.

\begin{center}
\begin{figure}[h!]
  \includegraphics[width=15cm,height=5cm]{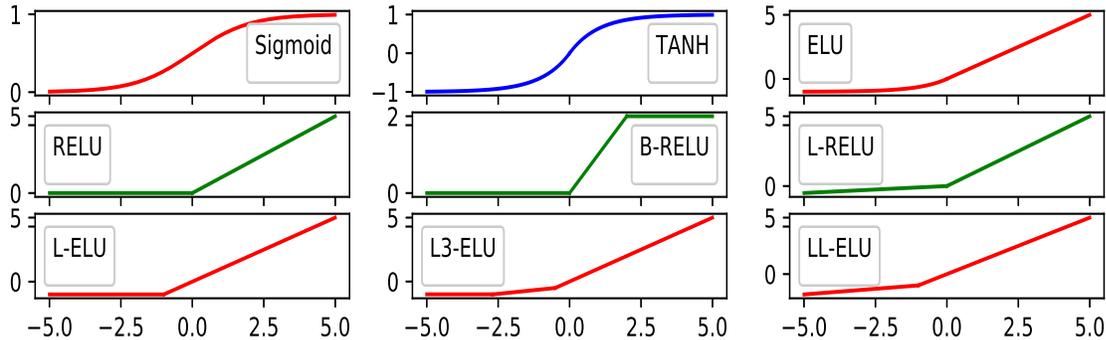}
  \label{fig:activ}
  \caption{Activation functions: sigmoid, hyperbolic tangent (TANH), Exponential Linear Unit (ELU), Rectified Linear Unit (RELU), Bounded-RELU (B-RELU), Leaky-RELU (L-RELU), Linearized-ELU (L-ELU), Linearized ELU with 3 segments (L3-ELU), Leaky Linearized ELU (LL-ELU). The first three are differentiable, the others are piecewise linear.}
\end{figure}
\end{center}

Finally we tried in each case a number of initial learning rates,
but we did not experiment with learning rate decay strategies
because there are too many possibilities, and because one could
expect that tampering with the learning rate $\eta$ may {\it a priori}
negatively interfere with the internal mechanics of the optimization
of the gradient  $\nabla$, since {\it in fine} both methods affect
the learning step $- \eta \nabla$. 

\section {Implementation and tests}

The algorithm was implemented as a short python program $xor.py$
described below using the TensorFlow library and 
derived from the program $neural\_network\_raw.py$
in the third section of TensorFlow-Examples of Aymeric Damien \cite{[19]}
To the credit of TensorFlow, the program runs as well on Biowulf,
the high performance super computer of the
NIH using GPUs, on a recent Mac, and on our 10-years old Ubuntu ThinkPad
laptop, although the GPU hardware is of course much faster when $p$ becomes large.

The training data is constructed by selecting at random successive 
batches of $10 p^2$
examples then adding 0.1 Gaussian noise to each input vector.
The program is trained on each batch, the training accuracy is 
evaluated, and the program stops when 
it reaches perfect accuracy on $20 p^2$ examples in a row.
The network is then evaluated
on a complete configuration of all $p^2$ couples $(a,b)$ without noise.
Randomness comes from the Gaussian initialization of the weights and
from the Gaussian noise in the inputs.
Each triplet (optimizer, activation, learning rate) was tested 10 times,
and success is reported when the network reaches
perfect accuracy on the test set. Since a number of methods
converged below 3000 epochs, the calculation is stopped if
it reaches 10,000 epoch and a FAILURE is reported. Analyzing
the dynamics of the training, one may distinguish 3 types of failures.
In some cases, the training accuracy never exceeds 0.40 or sometimes
goes a little higher then drops back irreversibly. In other cases,
the training seems to proceed well and relatively soon reaches a good
value over 0.85 or even over 0.98 for larger $p$, but stays trapped
for the remaining thousands of epochs in this false minimum. Finally in some
cases, the network reaches perfect accuracy  $20 p^2$ times in a row on 
the training set but fails on some configurations of the test set.

The first observation is that the problem seems much harder than
one could naively anticipate. The minimal number of epochs needed to
train the network is surprisingly large. 
Recalling that an epoch contains $10 p^2$ examples, and that the network is defined by
$3 p^2 + 2 p$ weights, we observe that the number of weight-updates needed
to reach perfect accuracy scales like $p^4$. 
Furthermore, in a great
number of cases, the network diverges or is trapped in a false
minimum (the many zeroes in table 3). 

\begin{center}
\begin{figure}[h!]
  \includegraphics[width=15cm,height=5cm]{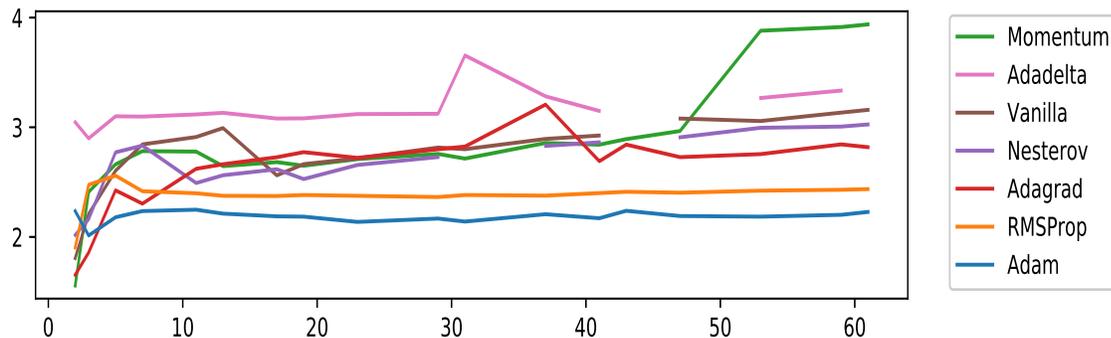}
  \label{fig:activ}
  \caption{$Log_{10}$ of the number of epochs of $10 p^2$ examples needed to perfectly train the network, using ELU and the optimal learning rate, for each optimizer.
 A gap indicates a case where less than 5 
trials out of 10 converged in less than 10000 epochs. A flat line indicates that the number of weight updates scales like $p^4$. For $p > 3$, Adam performs best, consistently 4 times faster than Adagrad, 10 times faster than Adadelta. 
}
\end{figure}
\end{center}

  \begin {table}
  \label{table:1}
\begin{tiny}
  \begin{tabular}{ ||c||c|c|c|c|c|c|c|c|c|c|c|c|c|c|c|c|c|c||}
    \hline
    \hline
	Classes & 2 & 3 & 5 & 7 & 11 & 13 & 17 & 19 & 23 & 29 & 31 & 37 & 41 & 43 & 47 & 53 & 59 & 61 \\
\hline
	adam & 172 & 103 & 151 & 172 & 177 & 163 & 154 & 153 & 137 & 147 & 138 & 161 & 148 & 173 & 155 & 153 & 159 & 169 \\
	RMSProp & 80 & 299 & 362 & 261 & 250 & 237 & 236 & 241 & 237 & 231 & 241 & 238 & 251 & 258 & 253 & 264 & 269 & 273 \\
	momentum & 36 & 257 & 459 & 606 & 600 & 442 & 481 & 446 & 512 & 571 & 518 & 719 & 694 & 780 & 924 & 7592 & 8202 & 8682 \\
	adagrad & 45 & 72 & 266 & 201 & 418 & 460 & 533 & 593 & 526 & 630 & 667 & 1611 & 490 & 695 & 534 & 569 & 698 & 658 \\
	Nesterov & 104 & 145 & 591 & 680 & 310 & 365 & 414 & 337 & 453 & 535 & 0 & 676 & 729 & 0 & 810 & 989 & 1015 & 1061 \\
	vanilla & 64 & 162 & 401 & 698 & 815 & 984 & 364 & 461 & 520 & 653 & 631 & 783 & 841 & 0 & 1201 & 1140 & 1361 & 1442 \\
	adadelta & 1112 & 791 & 1259 & 1252 & 1310 & 1354 & 1202 & 1206 & 1319 & 1327 & 4524 & 1915 & 1410 & 0 & 0 & 1849 & 2162 & 0 \\
\hline
    \hline
  \end {tabular}
\end{tiny}
  \caption {Lowest number of epochs of $10 p^2$ examples needed to solve exactly the 
generalized XOR classification problem 
with p classes, using ELU and optimal learning rate, counted on at least 5 successful trials. 
A zero indicates a case where less than 5 trials out of 10 converged in less than 10000 epochs.}
  \end {table}

Using ELU, and selecting in each experiment the optimal learning rate, we show in figure 3 (data in table 1)
the number of epochs, as a function of the number of classes,
needed to obtain a perfect network for each method. The zeroes denote cases
where less than half of the tests converged in less than
10,000 epochs, i.e. where the network was often trapped in a false minimum
or on a flat plateau. The Adam method is the most economic and most stable method
for all values of $p$, except  $p = 2$ where several methods converges faster. 
In non systematic tests up to 191, Adam remains the best method, followed by RMSprop.

The second  observation, figure 4 (data in table 2), is that for the Adam optimizer, the ELU activation function,
a combination of an exponential for negative x and the identity
for positive x (figure 2c), and its piece-wise linear approximations L3ELU, LLLU and LELU,
are preferable, for all
optimization methods, to the RELU, leaky-RELU and sigmoid functions. 
Table 3 at the end of the paper
in which the detailed results are presented for all prime numbers
$p <= 47$, and all activation functions, show that ELU is also preferable
for all the other optimizers that we tested. ELU is known to
be the best activation function, 
see for example reference 3, page 283, 
but it is not used so often in the literature, 
thus the striking superiority of ELU and its variants
over sigmoid and over RELU and its variants in this benchmark 
still comes as a surprise. 

A third observation is that contrary to the assertion of its creators \cite{[14]},
the differentiablility of the ELU function does not seem to play an important role.
ELU and L3ELU obtain very similar results. Actually, this is not really surprising
since computers always deal with finite differences rather than partial derivatives.
The advantage of ELU over RELU is more probably due to the fact that small
negative values, larger than -1, sieve through the ELU filter but are
blocked by the RELU, making the training of the network harder
as soon as a neuron gives a slightly negative results.
This is confirmed by the better results of the hyperbolic tangent over the sigmoid activation functions
although both are differentiable (Figure 2a and 2b). 

Notice that we did not see any difference of speed when using on our GPU
the exponential ELU activation relative to its linearized approximations.

\begin{center}
\begin{figure}[h!]
  \includegraphics[width=15cm,height=5cm]{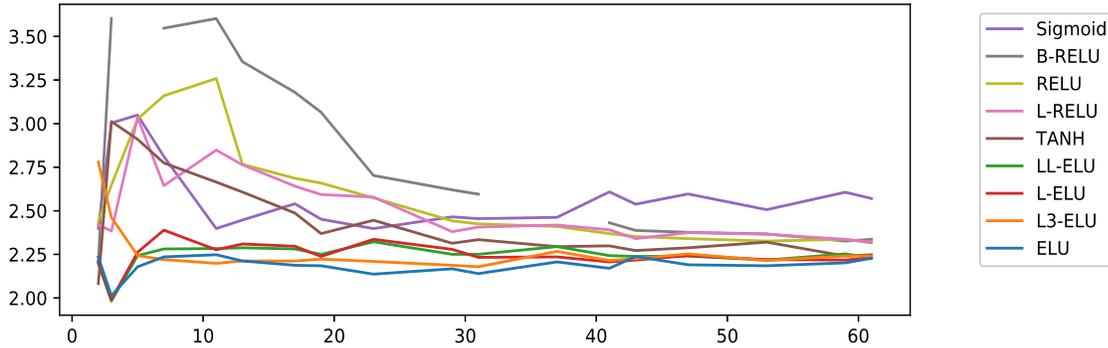}
  \label{fig:activ}
  \caption{$Log_{10}$ of the averaged number of epochs needed to perfectly train the network, using the Adam optimizer and the optimal learning rate, for each activation function. A gap indicates a case where less than 5 
trials out of 10 converged in less than 10000 epochs. The differentiability of the activation function is irrelevant: ELU and its linear approximations perform best. A flat line indicates that the number of weight updates scales like $p^4$.
}
\end{figure}
\end{center}

  \begin {table}
  \label{table:2}
\begin{tiny}
  \begin{tabular}{ ||c||c|c|c|c|c|c|c|c|c|c|c|c|c|c|c|c|c|c||}
    \hline
    \hline
	Classes & 2 & 3 & 5 & 7 & 11 & 13 & 17 & 19 & 23 & 29 & 31 & 37 & 41 & 43 & 47 & 53 & 59 & 61 \\
\hline
	ELU & 172 & 103 & 151 & 172 & 177 & 163 & 154 & 153 & 137 & 147 & 138 & 161 & 148 & 173 & 155 & 153 & 159 & 169 \\
	L3ELU & 602 & 290 & 175 & 166 & 158 & 163 & 163 & 167 & 162 & 154 & 151 & 185 & 164 & 168 & 179 & 164 & 174 & 174 \\
	LLELU & 157 & 96 & 175 & 191 & 192 & 194 & 191 & 178 & 210 & 178 & 178 & 197 & 175 & 173 & 174 & 165 & 179 & 170 \\
	LELU & 162 & 98 & 183 & 245 & 189 & 204 & 198 & 173 & 217 & 190 & 171 & 172 & 161 & 165 & 175 & 166 & 165 & 170 \\
	sigmoid & 251 & 1008 & 1120 & 648 & 250 & 280 & 347 & 283 & 250 & 292 & 285 & 290 & 406 & 345 & 395 & 321 & 404 & 372 \\
	TANH & 121 & 1028 & 813 & 595 & 462 & 405 & 307 & 234 & 279 & 206 & 216 & 197 & 199 & 187 & 194 & 209 & 174 & 177 \\
	L-R & 263 & 242 & 1084 & 441 & 705 & 580 & 438 & 392 & 379 & 240 & 255 & 262 & 246 & 219 & 237 & 232 & 217 & 209 \\
	RELU & 274 & 444 & 1064 & 1444 & 1814 & 584 & 486 & 456 & 375 & 277 & 266 & 257 & 234 & 225 & 219 & 212 & 217 & 207 \\
	B-R & 161 & 4010 & 0 & 3525 & 4003 & 2265 & 1512 & 1162 & 504 & 417 & 394 & 0 & 270 & 244 & 238 & 233 & 212 & 217 \\
\hline
    \hline
  \end {tabular}
\end{tiny}
  \caption {Averaged number of epochs in successful runs for each activation function using Adam 
optimization and the optimal learning rate. A zero indicates a case where less than 5 
trials out of 10 converged in less than 10000 epochs.}
  \end {table}

Finally, the optimal learning rate depends on the optimization method. The best 
results are obtained for a low learning rate, 0.1 or 0.01 in a few cases, for 
Adam and RMSprop. But for Adagrad, Adadelta, Nesterov
and Vanilla, a learning rate of 1 is preferable and even 5 when the
number of classes becomes larger (see table 3).

In a non exhaustive way, since the number of possibilities is open, we tried
other network designs, either with more or less than p cells in the hidden layer,
or with 2 or 3 hidden layers, always keeping $p$ cells in the output
layer which exports a $p$ dimensional 1-hot vector, using the
softmax activation function. Some of the alternative designs converged, for example $p=41$,
with only 20 hidden cells, or with 60 cells, but none converged faster
when timed on a CPU or when counting the number of weight updates.


\section{Using smaller data batches.}

We then tried smaller batch sizes. Downgrading from $10 p^2$ to $p^2$ 
training examples in each batch essentially 
gave the same results. Using $max(1,p^2/100)$ was too cahotic 
and the networks converged less frequently.
However, using $max(1,p^2/10)$ was interesting. The batches beeing a hundred 
times smaller, they run 100 times faster,
but because the data are more cahotic, the best learning rate for
this method is usually lower and the number of batches needed to reach 
a perfect accuracy about 10 times higher. As a result the program is around
10 times faster. This allowed us to test a few points up to $p=191$.
But the important lesson is that the relative performances of the optimizers 
and of the activation functions do not depend on the batch size.

The Adam optimizer, coupled to one of the variants of the ELU activation 
function remains best. RMS-prop also deals well with the smaller batches. 
The other results are consistent with the observations using the larger 
batches, with a notable exception: the new test is very harsh for the L4 optimizers. 

The principle of the  L4 method is to accelerate the convergence by constantly 
adapating the learning rate in the hope of lowering the loss at a constant 
exponential rate. Since our loss function is positive definite, convex and 
reaches zero at its minima, we are exactly in the condition of \cite{[13]}. 
L4-Adam performed very well on the large batches, independently of the 
initial learning rate. However, on the smaller 
batches, overtraining is manifest. The spikes of high learning 
rate overadapt to the peculiarites of the mini-batches which are now
too small to span the full landscape and L4-Adam becomes one of the 
very worst optimizers.


\section{Conclusion}

In this simple study, we have presented a natural generalization
of the seminal XOR problem to the classification of $p$-classes
using subtraction modulo $p$. Training and test examples are easy to
generate, and if $p$ is prime, the problem is indecomposable.
This problem is interesting because it can be solved with a single 
fully connected hidden layer, yet the parameter landscape is complex,
with a very high symmetry group and different classes of solutions.
As a result, the gradient descent is challenging and provides
a rich benchmark for activation functions, regulators and optimizers. 
We ran systematic tests up to $p = 61$, and a few up to $p = 191$.

We observed that the ELU activation function is preferable in all cases 
over the sigmoid and the RELU, and that the Adam optimizer converges
 more often and faster than, in order of performance, 
the RMSprop, Adagrad, Nesterov, Vanilla, Adatadelta and Momentum methods.
The L4 method, which fairs well on large batches, is subject to overtraining
on smaller batches.

Any better solution would be welcome, so would any design allowing to reach
the largest possible value of $p$, while maintaining perfect accuracy.


\section*{Source code}

The source code to reproduce on average table 1
is available from NCBI at

https://www.aceview.org/Software/Neural\_nets/XOR\_p/xor.py

The script is self-contained. It only depends on the standard TensorFlow
interface, and on the L4 optimizer downloaded as is from GitHub.
The top of the script explains how to duplicate the results presented here.
This python code is expected to run on any machine
where TensorFlow is available. Since there is an element of randomness in
the initialization, the table cannot be reproduced exactly, but
the conclusions should not change. The command line allows 
to specify the optimizer, the activation function, the learning 
rate, the batch size and the value of $p$. 
Try 'python xor.py --help' for details. 


\section*{Acknowledgments}

We would like to thank Mehmet Kayaalp at the NLM and Yann Thierry-Mieg at LIP6 for insightful suggestions,
our colleagues at the NCBI for numerous discussions on neural networks,
and especially David Landsman and Lewis Geer for actively promoting research on neural networks.
This research was supported by the Intramural Research Program of the NIH, U.S. National Library of Medicine.
This work utilized the computational resources of the NCBI and of the NIH HPC Biowulf cluster. (http://hpc.nih.gov)
and a 10 years old ThinkPad laptop.


\begin{addmargin}[1em]{1em}
  \begin {table}
  \label{table:3a}
\begin{tiny}
  \begin{tabular}{ ||c|c|c||c|c|c|c|c|c|c|c|c|c|c|c|c|c|c||}
    \hline
    \hline
	Method & Activation & Rate & 2 & 3 & 5 & 7 & 11 & 13 & 17 & 19 & 23 & 29 & 31 & 37 & 41 & 43 & 47 \\
\hline
\hline
	adam & ELU & 0.01 & 1207 & 535 & 733 & 709 & 785 & 811 & 771 & 769 & 786 & 834 & 769 & 848 & 827 & 0 & 810 \\
	adam & ELU & 0.1 & 119 & 209 & 219 & 186 & 177 & 163 & 154 & 153 & 137 & 147 & 138 & 161 & 148 & 151 & 155 \\
	adam & ELU & 1 & 3138 & 2613 & 2428 & 681 & 439 & 367 & 327 & 348 & 444 & 472 & 396 & 748 & 328 & 0 & 359 \\
	adam & ELU & 5 & 0 & 0 & 0 & 0 & 0 & 0 & 0 & 0 & 0 & 0 & 0 & 0 & 0 & 0 & 0 \\
	adam & sigmoid & 0.01 & 866 & 2306 & 1968 & 2466 & 2213 & 3938 & 2579 & 2194 & 0 & 0 & 2158 & 0 & 0 & 0 & 1986 \\
	adam & sigmoid & 0.1 & 305 & 471 & 937 & 595 & 569 & 762 & 451 & 716 & 621 & 409 & 483 & 440 & 491 & 555 & 409 \\
	adam & sigmoid & 1 & 2064 & 813 & 1794 & 1247 & 250 & 280 & 347 & 283 & 250 & 292 & 285 & 290 & 406 & 327 & 395 \\
	adam & sigmoid & 5 & 1773 & 0 & 0 & 0 & 4872 & 4587 & 0 & 0 & 0 & 0 & 0 & 0 & 0 & 0 & 0 \\
	adam & TANH & 0.01 & 0 & 2904 & 2502 & 1744 & 1791 & 1821 & 1413 & 1366 & 1299 & 1016 & 896 & 1103 & 866 & 0 & 981 \\
	adam & TANH & 0.1 & 0 & 604 & 551 & 384 & 462 & 408 & 307 & 234 & 279 & 223 & 259 & 252 & 223 & 186 & 195 \\
	adam & TANH & 1 & 2314 & 0 & 3461 & 1114 & 531 & 405 & 309 & 309 & 297 & 206 & 216 & 197 & 199 & 187 & 194 \\
	adam & TANH & 5 & 3009 & 0 & 0 & 0 & 0 & 0 & 0 & 0 & 0 & 0 & 0 & 0 & 0 & 0 & 0 \\
	adam & L-R & 0.01 & 505 & 1291 & 1845 & 1849 & 2302 & 1592 & 1508 & 1384 & 1051 & 975 & 1108 & 1031 & 945 & 1202 & 873 \\
	adam & L-R & 0.1 & 122 & 387 & 575 & 1648 & 705 & 580 & 438 & 392 & 379 & 240 & 255 & 262 & 246 & 216 & 237 \\
	adam & L-R & 1 & 2369 & 1155 & 0 & 2926 & 1989 & 792 & 468 & 950 & 1594 & 680 & 466 & 486 & 959 & 1673 & 1984 \\
	adam & L-R & 5 & 0 & 0 & 0 & 0 & 0 & 0 & 0 & 0 & 0 & 0 & 0 & 0 & 0 & 0 & 0 \\
	adam & RELU & 0.01 & 1569 & 1292 & 1533 & 2462 & 1814 & 1788 & 1525 & 1307 & 1232 & 1085 & 1023 & 1075 & 873 & 1137 & 940 \\
	adam & RELU & 0.1 & 0 & 2772 & 4334 & 1823 & 2077 & 584 & 486 & 456 & 375 & 277 & 266 & 257 & 234 & 240 & 219 \\
	adam & RELU & 1 & 0 & 0 & 0 & 0 & 0 & 0 & 0 & 0 & 0 & 0 & 0 & 0 & 0 & 0 & 0 \\
	adam & RELU & 5 & 0 & 0 & 0 & 0 & 0 & 0 & 0 & 0 & 0 & 0 & 0 & 0 & 0 & 0 & 0 \\
	adam & B-R & 0.01 & 600 & 0 & 2666 & 0 & 4003 & 4009 & 2339 & 2273 & 1993 & 1445 & 1716 & 0 & 1164 & 1168 & 1095 \\
	adam & B-R & 0.1 & 2868 & 1573 & 0 & 0 & 0 & 2265 & 1512 & 1162 & 504 & 417 & 394 & 0 & 270 & 0 & 238 \\
	adam & B-R & 1 & 0 & 0 & 0 & 0 & 0 & 0 & 0 & 0 & 0 & 0 & 0 & 0 & 0 & 0 & 0 \\
	adam & B-R & 5 & 0 & 0 & 0 & 0 & 0 & 0 & 0 & 0 & 0 & 0 & 0 & 0 & 0 & 0 & 0 \\
	adam & LELU & 0.01 & 248 & 537 & 1056 & 1268 & 1097 & 1049 & 1142 & 1225 & 1280 & 1000 & 976 & 1447 & 865 & 0 & 996 \\
	adam & LELU & 0.1 & 97 & 110 & 168 & 246 & 189 & 204 & 198 & 173 & 217 & 190 & 171 & 172 & 161 & 147 & 175 \\
	adam & LELU & 1 & 0 & 0 & 0 & 0 & 2794 & 1776 & 2547 & 2579 & 2293 & 0 & 0 & 0 & 0 & 0 & 0 \\
	adam & LELU & 5 & 0 & 0 & 0 & 0 & 0 & 0 & 0 & 0 & 0 & 0 & 0 & 0 & 0 & 0 & 0 \\
	adam & LLELU & 0.01 & 200 & 661 & 1150 & 1020 & 1082 & 1540 & 1231 & 1206 & 0 & 1081 & 1144 & 1125 & 0 & 0 & 1039 \\
	adam & LLELU & 0.1 & 916 & 121 & 189 & 147 & 192 & 194 & 191 & 178 & 210 & 178 & 185 & 197 & 175 & 0 & 174 \\
	adam & LLELU & 1 & 326 & 380 & 260 & 402 & 488 & 445 & 277 & 231 & 273 & 203 & 178 & 0 & 0 & 0 & 178 \\
	adam & LLELU & 5 & 449 & 641 & 1062 & 767 & 398 & 421 & 0 & 0 & 0 & 0 & 0 & 0 & 0 & 0 & 0 \\
	adam & L3ELU & 0.01 & 206 & 486 & 780 & 978 & 1018 & 924 & 929 & 1062 & 1060 & 826 & 939 & 1069 & 884 & 0 & 856 \\
	adam & L3ELU & 0.1 & 1003 & 104 & 148 & 136 & 158 & 163 & 163 & 167 & 162 & 154 & 151 & 185 & 164 & 169 & 179 \\
	adam & L3ELU & 1 & 0 & 0 & 0 & 1827 & 627 & 571 & 496 & 520 & 969 & 5161 & 0 & 0 & 566 & 0 & 430 \\
	adam & L3ELU & 5 & 0 & 0 & 0 & 0 & 0 & 0 & 0 & 0 & 0 & 0 & 0 & 0 & 0 & 0 & 0 \\
\hline
	RMSProp & ELU & 0.01 & 579 & 281 & 415 & 500 & 470 & 490 & 479 & 481 & 465 & 454 & 467 & 483 & 481 & 468 & 482 \\
	RMSProp & ELU & 0.1 & 77 & 190 & 268 & 253 & 250 & 237 & 236 & 241 & 237 & 231 & 241 & 238 & 251 & 254 & 253 \\
	RMSProp & ELU & 1 & 75 & 2350 & 3736 & 0 & 0 & 0 & 1057 & 449 & 0 & 0 & 0 & 0 & 0 & 0 & 0 \\
	RMSProp & ELU & 5 & 0 & 0 & 0 & 0 & 0 & 0 & 0 & 0 & 0 & 0 & 0 & 0 & 0 & 0 & 0 \\
	RMSProp & sigmoid & 0.01 & 183 & 1236 & 1511 & 1460 & 2100 & 1667 & 1495 & 2407 & 1260 & 1481 & 1262 & 2189 & 1771 & 1148 & 1335 \\
	RMSProp & sigmoid & 0.1 & 92 & 897 & 1729 & 922 & 2081 & 0 & 6520 & 6078 & 9021 & 7993 & 0 & 0 & 10000 & 10000 & 10000 \\
	RMSProp & sigmoid & 1 & 61 & 0 & 0 & 0 & 0 & 0 & 0 & 0 & 0 & 0 & 0 & 0 & 0 & 0 & 0 \\
	RMSProp & sigmoid & 5 & 0 & 0 & 0 & 0 & 0 & 0 & 0 & 0 & 0 & 0 & 0 & 0 & 0 & 0 & 0 \\
	RMSProp & TANH & 0.01 & 196 & 2074 & 2576 & 1371 & 747 & 972 & 732 & 677 & 649 & 587 & 656 & 745 & 620 & 655 & 627 \\
	RMSProp & TANH & 0.1 & 59 & 545 & 1091 & 683 & 1015 & 757 & 488 & 441 & 501 & 386 & 368 & 395 & 334 & 407 & 358 \\
	RMSProp & TANH & 1 & 2357 & 3500 & 3698 & 4408 & 4355 & 3545 & 4317 & 5709 & 0 & 0 & 0 & 0 & 0 & 0 & 0 \\
	RMSProp & TANH & 5 & 3038 & 0 & 0 & 0 & 0 & 0 & 0 & 0 & 0 & 0 & 0 & 0 & 0 & 0 & 0 \\
	RMSProp & L-R & 0.01 & 719 & 695 & 895 & 1735 & 1206 & 1296 & 1020 & 888 & 672 & 584 & 586 & 574 & 502 & 505 & 486 \\
	RMSProp & L-R & 0.1 & 108 & 197 & 570 & 738 & 934 & 972 & 578 & 453 & 431 & 350 & 348 & 372 & 343 & 319 & 323 \\
	RMSProp & L-R & 1 & 414 & 0 & 0 & 0 & 0 & 0 & 0 & 0 & 0 & 0 & 0 & 0 & 0 & 0 & 0 \\
	RMSProp & L-R & 5 & 2141 & 0 & 0 & 0 & 0 & 0 & 0 & 0 & 0 & 0 & 0 & 0 & 0 & 0 & 0 \\
	RMSProp & RELU & 0.01 & 552 & 0 & 1300 & 1645 & 1101 & 1062 & 1129 & 715 & 730 & 602 & 561 & 528 & 491 & 504 & 503 \\
	RMSProp & RELU & 0.1 & 126 & 0 & 1359 & 2668 & 1067 & 1624 & 600 & 466 & 480 & 430 & 375 & 353 & 336 & 371 & 364 \\
	RMSProp & RELU & 1 & 0 & 0 & 0 & 0 & 0 & 0 & 0 & 0 & 0 & 0 & 0 & 0 & 0 & 0 & 0 \\
	RMSProp & RELU & 5 & 0 & 0 & 0 & 0 & 0 & 0 & 0 & 0 & 0 & 0 & 0 & 0 & 0 & 0 & 0 \\
	RMSProp & B-R & 0.01 & 0 & 0 & 0 & 1740 & 5212 & 0 & 1658 & 1654 & 1489 & 754 & 877 & 791 & 661 & 666 & 672 \\
	RMSProp & B-R & 0.1 & 0 & 0 & 0 & 0 & 0 & 0 & 2074 & 1133 & 1328 & 632 & 551 & 516 & 430 & 513 & 439 \\
	RMSProp & B-R & 1 & 0 & 0 & 0 & 0 & 0 & 0 & 0 & 0 & 0 & 0 & 0 & 0 & 0 & 0 & 0 \\
	RMSProp & B-R & 5 & 0 & 0 & 0 & 0 & 0 & 0 & 0 & 0 & 0 & 0 & 0 & 0 & 0 & 0 & 0 \\
	RMSProp & LELU & 0.01 & 194 & 443 & 503 & 614 & 741 & 774 & 617 & 697 & 650 & 577 & 613 & 605 & 577 & 600 & 606 \\
	RMSProp & LELU & 0.1 & 212 & 665 & 330 & 407 & 283 & 333 & 299 & 282 & 285 & 272 & 286 & 298 & 313 & 288 & 311 \\
	RMSProp & LELU & 1 & 1770 & 0 & 0 & 0 & 0 & 0 & 0 & 0 & 0 & 0 & 0 & 0 & 0 & 0 & 0 \\
	RMSProp & LELU & 5 & 0 & 0 & 0 & 0 & 0 & 0 & 0 & 0 & 0 & 0 & 0 & 0 & 0 & 0 & 0 \\
	RMSProp & LLELU & 0.01 & 286 & 340 & 591 & 782 & 706 & 955 & 938 & 757 & 782 & 824 & 706 & 726 & 703 & 725 & 689 \\
	RMSProp & LLELU & 0.1 & 145 & 174 & 361 & 370 & 358 & 374 & 359 & 310 & 310 & 300 & 283 & 287 & 291 & 298 & 304 \\
	RMSProp & LLELU & 1 & 165 & 0 & 0 & 0 & 0 & 0 & 490 & 323 & 264 & 241 & 218 & 0 & 0 & 0 & 0 \\
	RMSProp & LLELU & 5 & 0 & 0 & 0 & 0 & 0 & 0 & 0 & 0 & 0 & 0 & 0 & 0 & 0 & 0 & 0 \\
	RMSProp & L3ELU & 0.01 & 269 & 291 & 574 & 635 & 656 & 570 & 599 & 559 & 556 & 527 & 535 & 568 & 548 & 543 & 572 \\
	RMSProp & L3ELU & 0.1 & 131 & 192 & 293 & 271 & 303 & 252 & 261 & 265 & 249 & 264 & 255 & 263 & 270 & 285 & 280 \\
	RMSProp & L3ELU & 1 & 0 & 0 & 0 & 0 & 0 & 0 & 0 & 0 & 0 & 0 & 0 & 0 & 0 & 0 & 0 \\
	RMSProp & L3ELU & 5 & 0 & 0 & 0 & 0 & 0 & 0 & 0 & 0 & 0 & 0 & 0 & 0 & 0 & 0 & 0 \\
\hline
    \hline
  \end {tabular}
\end{tiny}
  \end {table}

  \begin {table}
  \label{table:3b}
\begin{tiny}
  \begin{tabular}{ ||c|c|c||c|c|c|c|c|c|c|c|c|c|c|c|c|c|c||}
    \hline
    \hline
	Method & Activation & Rate & 2 & 3 & 5 & 7 & 11 & 13 & 17 & 19 & 23 & 29 & 31 & 37 & 41 & 43 & 47 \\
\hline
    \hline
	momentum & ELU & 0.01 & 328 & 1087 & 4244 & 4762 & 0 & 0 & 0 & 0 & 0 & 0 & 0 & 0 & 0 & 0 & 0 \\
	momentum & ELU & 0.1 & 242 & 308 & 1365 & 691 & 1142 & 1587 & 1873 & 2253 & 3156 & 3317 & 3348 & 0 & 5293 & 0 & 6219 \\
	momentum & ELU & 1 & 33 & 519 & 462 & 519 & 600 & 442 & 481 & 446 & 512 & 571 & 518 & 719 & 694 & 0 & 924 \\
	momentum & ELU & 5 & 0 & 0 & 0 & 0 & 0 & 0 & 0 & 0 & 0 & 0 & 0 & 0 & 0 & 0 & 0 \\
	momentum & sigmoid & 0.01 & 977 & 5683 & 0 & 0 & 0 & 0 & 0 & 0 & 0 & 0 & 0 & 0 & 0 & 0 & 0 \\
	momentum & sigmoid & 0.1 & 0 & 2289 & 2622 & 4344 & 6952 & 0 & 0 & 0 & 0 & 0 & 0 & 0 & 0 & 0 & 0 \\
	momentum & sigmoid & 1 & 194 & 287 & 1574 & 2803 & 1929 & 2870 & 1238 & 0 & 1354 & 1943 & 2020 & 3462 & 2317 & 2423 & 3525 \\
	momentum & sigmoid & 5 & 212 & 394 & 760 & 621 & 603 & 590 & 541 & 515 & 0 & 685 & 769 & 816 & 839 & 859 & 901 \\
	momentum & TANH & 0.01 & 252 & 0 & 0 & 0 & 0 & 0 & 0 & 0 & 0 & 0 & 0 & 0 & 0 & 0 & 0 \\
	momentum & TANH & 0.1 & 60 & 1783 & 2725 & 4567 & 3593 & 4044 & 5411 & 5242 & 5384 & 6720 & 0 & 7665 & 6656 & 0 & 0 \\
	momentum & TANH & 1 & 396 & 1115 & 1580 & 1623 & 1200 & 1270 & 720 & 734 & 632 & 938 & 749 & 995 & 951 & 1100 & 1026 \\
	momentum & TANH & 5 & 1396 & 0 & 0 & 0 & 0 & 0 & 0 & 0 & 0 & 0 & 0 & 0 & 0 & 0 & 0 \\
	momentum & L-R & 0.01 & 0 & 0 & 0 & 0 & 0 & 0 & 0 & 0 & 0 & 0 & 0 & 0 & 0 & 0 & 0 \\
	momentum & L-R & 0.1 & 2512 & 4333 & 3909 & 0 & 0 & 0 & 0 & 0 & 0 & 0 & 0 & 0 & 0 & 0 & 0 \\
	momentum & L-R & 1 & 1087 & 711 & 1075 & 2691 & 4854 & 3659 & 4404 & 5097 & 5359 & 0 & 0 & 0 & 0 & 0 & 0 \\
	momentum & L-R & 5 & 0 & 0 & 0 & 0 & 1795 & 1508 & 1088 & 1274 & 1465 & 2866 & 4555 & 0 & 0 & 0 & 0 \\
	momentum & RELU & 0.01 & 0 & 0 & 0 & 0 & 0 & 0 & 0 & 0 & 0 & 0 & 0 & 0 & 0 & 0 & 0 \\
	momentum & RELU & 0.1 & 49 & 0 & 0 & 0 & 0 & 0 & 0 & 0 & 0 & 0 & 0 & 0 & 0 & 0 & 0 \\
	momentum & RELU & 1 & 0 & 0 & 0 & 0 & 0 & 0 & 0 & 0 & 0 & 0 & 0 & 0 & 0 & 0 & 0 \\
	momentum & RELU & 5 & 0 & 0 & 0 & 0 & 0 & 0 & 0 & 0 & 0 & 0 & 0 & 0 & 0 & 0 & 0 \\
	momentum & B-R & 0.01 & 0 & 0 & 0 & 0 & 0 & 0 & 0 & 0 & 0 & 0 & 0 & 0 & 0 & 0 & 0 \\
	momentum & B-R & 0.1 & 0 & 0 & 0 & 0 & 0 & 0 & 0 & 0 & 0 & 0 & 0 & 0 & 0 & 0 & 0 \\
	momentum & B-R & 1 & 0 & 0 & 0 & 0 & 0 & 0 & 0 & 0 & 0 & 0 & 0 & 0 & 0 & 0 & 0 \\
	momentum & B-R & 5 & 0 & 0 & 0 & 0 & 0 & 0 & 0 & 0 & 0 & 0 & 0 & 0 & 0 & 0 & 0 \\
	momentum & LELU & 0.01 & 0 & 911 & 2643 & 7481 & 0 & 0 & 0 & 0 & 0 & 0 & 0 & 0 & 0 & 0 & 0 \\
	momentum & LELU & 0.1 & 0 & 1145 & 1062 & 1089 & 3690 & 1786 & 3517 & 3699 & 0 & 6706 & 6904 & 0 & 0 & 0 & 0 \\
	momentum & LELU & 1 & 632 & 1305 & 2475 & 2446 & 2957 & 2987 & 2443 & 2400 & 1960 & 1812 & 1701 & 1931 & 0 & 0 & 0 \\
	momentum & LELU & 5 & 0 & 0 & 0 & 0 & 0 & 0 & 0 & 0 & 0 & 0 & 0 & 0 & 0 & 0 & 0 \\
	momentum & LLELU & 0.01 & 671 & 1906 & 4204 & 6679 & 0 & 0 & 0 & 0 & 0 & 0 & 0 & 0 & 0 & 0 & 0 \\
	momentum & LLELU & 0.1 & 1332 & 372 & 468 & 887 & 2513 & 2296 & 0 & 2556 & 0 & 4910 & 5135 & 0 & 0 & 0 & 8205 \\
	momentum & LLELU & 1 & 747 & 379 & 189 & 790 & 454 & 850 & 412 & 421 & 0 & 0 & 0 & 0 & 0 & 0 & 0 \\
	momentum & LLELU & 5 & 0 & 0 & 0 & 0 & 0 & 0 & 0 & 0 & 0 & 0 & 0 & 0 & 0 & 0 & 0 \\
	momentum & L3ELU & 0.01 & 248 & 2238 & 3075 & 5842 & 0 & 0 & 0 & 0 & 0 & 0 & 0 & 0 & 0 & 0 & 0 \\
	momentum & L3ELU & 0.1 & 1051 & 173 & 402 & 754 & 1378 & 2100 & 1925 & 0 & 4029 & 5144 & 0 & 7697 & 8591 & 0 & 0 \\
	momentum & L3ELU & 1 & 30 & 2110 & 560 & 1299 & 601 & 758 & 952 & 613 & 796 & 969 & 886 & 0 & 0 & 1148 & 0 \\
	momentum & L3ELU & 5 & 0 & 0 & 0 & 0 & 0 & 0 & 0 & 0 & 0 & 0 & 0 & 0 & 0 & 0 & 0 \\
\hline
	adagrad & ELU & 0.01 & 3555 & 0 & 0 & 0 & 0 & 0 & 0 & 0 & 0 & 0 & 0 & 0 & 0 & 0 & 0 \\
	adagrad & ELU & 0.1 & 172 & 646 & 2640 & 3293 & 3431 & 3781 & 6297 & 5859 & 7988 & 0 & 0 & 0 & 0 & 0 & 0 \\
	adagrad & ELU & 1 & 46 & 910 & 273 & 213 & 418 & 460 & 533 & 593 & 773 & 1065 & 1159 & 1611 & 1576 & 0 & 1891 \\
	adagrad & ELU & 5 & 2124 & 3468 & 1770 & 1924 & 865 & 882 & 955 & 668 & 526 & 630 & 667 & 0 & 490 & 0 & 534 \\
	adagrad & sigmoid & 0.01 & 0 & 0 & 0 & 0 & 0 & 0 & 0 & 0 & 0 & 0 & 0 & 0 & 0 & 0 & 0 \\
	adagrad & sigmoid & 0.1 & 1278 & 2999 & 0 & 0 & 0 & 0 & 0 & 0 & 0 & 0 & 0 & 0 & 0 & 0 & 0 \\
	adagrad & sigmoid & 1 & 64 & 1863 & 2000 & 2914 & 2961 & 3365 & 0 & 0 & 0 & 0 & 6332 & 6984 & 0 & 0 & 6664 \\
	adagrad & sigmoid & 5 & 419 & 745 & 0 & 2210 & 1063 & 1622 & 1239 & 1145 & 1369 & 1196 & 1757 & 2187 & 1595 & 3317 & 1557 \\
	adagrad & TANH & 0.01 & 0 & 0 & 0 & 0 & 0 & 0 & 0 & 0 & 0 & 0 & 0 & 0 & 0 & 0 & 0 \\
	adagrad & TANH & 0.1 & 145 & 1226 & 0 & 0 & 0 & 0 & 0 & 0 & 0 & 0 & 0 & 0 & 0 & 0 & 0 \\
	adagrad & TANH & 1 & 0 & 1375 & 643 & 2897 & 1146 & 2022 & 1506 & 1558 & 2278 & 0 & 2172 & 2449 & 2476 & 3203 & 3175 \\
	adagrad & TANH & 5 & 0 & 5076 & 3776 & 3227 & 2715 & 2405 & 1035 & 1188 & 890 & 859 & 1178 & 0 & 1255 & 0 & 1447 \\
	adagrad & L-R & 0.01 & 0 & 0 & 0 & 0 & 0 & 0 & 0 & 0 & 0 & 0 & 0 & 0 & 0 & 0 & 0 \\
	adagrad & L-R & 0.1 & 948 & 3085 & 5563 & 0 & 0 & 0 & 0 & 0 & 0 & 0 & 0 & 0 & 0 & 0 & 0 \\
	adagrad & L-R & 1 & 795 & 2354 & 949 & 1883 & 3852 & 4071 & 5122 & 4264 & 4262 & 3776 & 3950 & 4224 & 4185 & 0 & 4656 \\
	adagrad & L-R & 5 & 0 & 0 & 0 & 4454 & 3514 & 3988 & 2716 & 4334 & 5307 & 4058 & 2860 & 2870 & 2132 & 1554 & 1669 \\
	adagrad & RELU & 0.01 & 0 & 0 & 0 & 0 & 0 & 0 & 0 & 0 & 0 & 0 & 0 & 0 & 0 & 0 & 0 \\
	adagrad & RELU & 0.1 & 0 & 0 & 3151 & 0 & 0 & 0 & 0 & 0 & 0 & 0 & 0 & 0 & 0 & 0 & 0 \\
	adagrad & RELU & 1 & 0 & 0 & 0 & 0 & 5842 & 4916 & 3572 & 4413 & 4875 & 4526 & 5359 & 5309 & 5028 & 5935 & 5779 \\
	adagrad & RELU & 5 & 0 & 0 & 0 & 0 & 0 & 0 & 0 & 0 & 0 & 6300 & 0 & 2421 & 2420 & 2022 & 2132 \\
	adagrad & B-R & 0.01 & 0 & 0 & 0 & 0 & 0 & 0 & 0 & 0 & 0 & 0 & 0 & 0 & 0 & 0 & 0 \\
	adagrad & B-R & 0.1 & 0 & 0 & 0 & 0 & 0 & 0 & 0 & 0 & 0 & 0 & 0 & 0 & 0 & 0 & 0 \\
	adagrad & B-R & 1 & 0 & 0 & 0 & 0 & 0 & 0 & 0 & 0 & 0 & 6756 & 0 & 0 & 6009 & 0 & 0 \\
	adagrad & B-R & 5 & 0 & 0 & 0 & 0 & 0 & 0 & 0 & 0 & 0 & 0 & 6240 & 3551 & 2279 & 2353 & 1707 \\
	adagrad & LELU & 0.01 & 0 & 0 & 0 & 0 & 0 & 0 & 0 & 0 & 0 & 0 & 0 & 0 & 0 & 0 & 0 \\
	adagrad & LELU & 0.1 & 0 & 651 & 2781 & 3165 & 5321 & 5653 & 7910 & 8975 & 0 & 0 & 0 & 0 & 0 & 0 & 0 \\
	adagrad & LELU & 1 & 1333 & 100 & 203 & 395 & 520 & 852 & 1033 & 920 & 1085 & 1367 & 1648 & 2020 & 2116 & 0 & 2871 \\
	adagrad & LELU & 5 & 0 & 0 & 0 & 0 & 6923 & 5791 & 5409 & 4390 & 4297 & 2758 & 0 & 1618 & 1608 & 0 & 1196 \\
	adagrad & LLELU & 0.01 & 2719 & 0 & 0 & 0 & 0 & 0 & 0 & 0 & 0 & 0 & 0 & 0 & 0 & 0 & 0 \\
	adagrad & LLELU & 0.1 & 91 & 733 & 1930 & 2841 & 5075 & 6211 & 8152 & 0 & 0 & 0 & 0 & 0 & 0 & 0 & 0 \\
	adagrad & LLELU & 1 & 67 & 160 & 349 & 323 & 498 & 591 & 710 & 864 & 0 & 1353 & 1547 & 2123 & 2481 & 0 & 2206 \\
	adagrad & LLELU & 5 & 1641 & 826 & 928 & 1155 & 780 & 659 & 582 & 848 & 740 & 597 & 695 & 765 & 784 & 0 & 653 \\
	adagrad & L3ELU & 0.01 & 0 & 0 & 0 & 0 & 0 & 0 & 0 & 0 & 0 & 0 & 0 & 0 & 0 & 0 & 0 \\
	adagrad & L3ELU & 0.1 & 263 & 842 & 1765 & 2637 & 4101 & 6697 & 6477 & 7551 & 0 & 0 & 0 & 0 & 0 & 0 & 0 \\
	adagrad & L3ELU & 1 & 37 & 103 & 195 & 273 & 412 & 533 & 652 & 678 & 1124 & 1328 & 1495 & 1822 & 2138 & 0 & 2609 \\
	adagrad & L3ELU & 5 & 0 & 0 & 0 & 3420 & 3483 & 1882 & 1637 & 0 & 825 & 980 & 715 & 934 & 1237 & 0 & 686 \\
\hline
    \hline
  \end {tabular}
\end{tiny}
  \end {table}

  \begin {table}
  \label{table:3c}
\begin{tiny}
  \begin{tabular}{ ||c|c|c||c|c|c|c|c|c|c|c|c|c|c|c|c|c|c||}
    \hline
    \hline
	Method & Activation & Rate & 2 & 3 & 5 & 7 & 11 & 13 & 17 & 19 & 23 & 29 & 31 & 37 & 41 & 43 & 47 \\
\hline
	Nesterov & ELU & 0.01 & 515 & 1208 & 2320 & 4691 & 0 & 0 & 0 & 0 & 0 & 0 & 0 & 0 & 0 & 0 & 0 \\
	Nesterov & ELU & 0.1 & 106 & 205 & 350 & 715 & 1212 & 1409 & 1789 & 2234 & 2448 & 3495 & 0 & 5360 & 0 & 0 & 0 \\
	Nesterov & ELU & 1 & 1043 & 1273 & 205 & 616 & 310 & 365 & 414 & 337 & 453 & 535 & 0 & 676 & 729 & 0 & 810 \\
	Nesterov & ELU & 5 & 0 & 0 & 0 & 0 & 0 & 0 & 0 & 0 & 0 & 0 & 0 & 0 & 0 & 0 & 0 \\
	Nesterov & sigmoid & 0.01 & 0 & 3247 & 0 & 0 & 0 & 0 & 0 & 0 & 0 & 0 & 0 & 0 & 0 & 0 & 0 \\
	Nesterov & sigmoid & 0.1 & 274 & 1197 & 2987 & 0 & 0 & 0 & 0 & 0 & 0 & 0 & 0 & 0 & 0 & 0 & 0 \\
	Nesterov & sigmoid & 1 & 43 & 967 & 572 & 2010 & 1679 & 5238 & 0 & 2465 & 3866 & 1698 & 3236 & 1862 & 2383 & 1921 & 3050 \\
	Nesterov & sigmoid & 5 & 662 & 1655 & 1650 & 966 & 726 & 845 & 609 & 529 & 552 & 624 & 661 & 747 & 700 & 807 & 815 \\
	Nesterov & TANH & 0.01 & 237 & 0 & 0 & 0 & 0 & 0 & 0 & 0 & 0 & 0 & 0 & 0 & 0 & 0 & 0 \\
	Nesterov & TANH & 0.1 & 0 & 1130 & 3479 & 3264 & 3929 & 3183 & 4458 & 4799 & 6681 & 0 & 6460 & 7343 & 7339 & 0 & 0 \\
	Nesterov & TANH & 1 & 56 & 949 & 1149 & 2390 & 774 & 966 & 581 & 998 & 790 & 773 & 848 & 2009 & 950 & 1102 & 938 \\
	Nesterov & TANH & 5 & 0 & 0 & 0 & 0 & 0 & 0 & 0 & 0 & 0 & 0 & 0 & 0 & 0 & 0 & 0 \\
	Nesterov & L-R & 0.01 & 0 & 0 & 0 & 0 & 0 & 0 & 0 & 0 & 0 & 0 & 0 & 0 & 0 & 0 & 0 \\
	Nesterov & L-R & 0.1 & 503 & 1718 & 4190 & 6993 & 0 & 0 & 0 & 0 & 0 & 0 & 0 & 0 & 0 & 0 & 0 \\
	Nesterov & L-R & 1 & 1800 & 1138 & 3400 & 1816 & 3071 & 3032 & 3227 & 4643 & 4660 & 6039 & 0 & 0 & 0 & 0 & 0 \\
	Nesterov & L-R & 5 & 0 & 0 & 0 & 0 & 0 & 0 & 0 & 0 & 0 & 0 & 0 & 0 & 0 & 0 & 0 \\
	Nesterov & RELU & 0.01 & 0 & 0 & 0 & 0 & 0 & 0 & 0 & 0 & 0 & 0 & 0 & 0 & 0 & 0 & 0 \\
	Nesterov & RELU & 0.1 & 0 & 0 & 0 & 0 & 0 & 0 & 0 & 0 & 0 & 0 & 0 & 0 & 0 & 0 & 0 \\
	Nesterov & RELU & 1 & 0 & 0 & 0 & 0 & 0 & 0 & 0 & 0 & 0 & 0 & 0 & 0 & 0 & 0 & 0 \\
	Nesterov & RELU & 5 & 0 & 0 & 0 & 0 & 0 & 0 & 0 & 0 & 0 & 0 & 0 & 0 & 0 & 0 & 0 \\
	Nesterov & B-R & 0.01 & 0 & 0 & 0 & 0 & 0 & 0 & 0 & 0 & 0 & 0 & 0 & 0 & 0 & 0 & 0 \\
	Nesterov & B-R & 0.1 & 0 & 0 & 0 & 0 & 0 & 0 & 0 & 0 & 0 & 0 & 0 & 0 & 0 & 0 & 0 \\
	Nesterov & B-R & 1 & 0 & 0 & 0 & 0 & 0 & 0 & 0 & 0 & 0 & 0 & 0 & 0 & 0 & 0 & 0 \\
	Nesterov & B-R & 5 & 0 & 0 & 0 & 0 & 0 & 0 & 0 & 0 & 0 & 0 & 0 & 0 & 0 & 0 & 0 \\
	Nesterov & LELU & 0.01 & 257 & 976 & 3087 & 4997 & 0 & 0 & 0 & 0 & 0 & 0 & 0 & 0 & 0 & 0 & 0 \\
	Nesterov & LELU & 0.1 & 1703 & 2051 & 1508 & 724 & 2051 & 2839 & 4491 & 0 & 0 & 6575 & 7000 & 0 & 0 & 0 & 0 \\
	Nesterov & LELU & 1 & 0 & 0 & 0 & 1986 & 917 & 1976 & 1460 & 1695 & 1298 & 1756 & 1666 & 1681 & 1635 & 1728 & 1785 \\
	Nesterov & LELU & 5 & 0 & 0 & 0 & 0 & 0 & 0 & 0 & 0 & 0 & 0 & 0 & 0 & 0 & 0 & 0 \\
	Nesterov & LLELU & 0.01 & 227 & 1329 & 4339 & 5545 & 0 & 0 & 0 & 0 & 0 & 0 & 0 & 0 & 0 & 0 & 0 \\
	Nesterov & LLELU & 0.1 & 133 & 176 & 625 & 672 & 1638 & 1870 & 0 & 3308 & 0 & 4711 & 5200 & 0 & 8023 & 0 & 8372 \\
	Nesterov & LLELU & 1 & 91 & 1139 & 316 & 287 & 418 & 336 & 748 & 575 & 738 & 602 & 0 & 0 & 0 & 0 & 0 \\
	Nesterov & LLELU & 5 & 0 & 0 & 0 & 0 & 0 & 0 & 0 & 0 & 0 & 0 & 0 & 0 & 0 & 0 & 0 \\
	Nesterov & L3ELU & 0.01 & 397 & 1362 & 4075 & 5790 & 0 & 0 & 0 & 0 & 0 & 0 & 0 & 0 & 0 & 0 & 0 \\
	Nesterov & L3ELU & 0.1 & 704 & 148 & 491 & 714 & 1475 & 1616 & 2151 & 2595 & 0 & 4632 & 0 & 0 & 0 & 0 & 0 \\
	Nesterov & L3ELU & 1 & 565 & 747 & 1105 & 585 & 745 & 975 & 628 & 635 & 895 & 851 & 913 & 1154 & 1122 & 0 & 0 \\
	Nesterov & L3ELU & 5 & 0 & 0 & 0 & 0 & 0 & 0 & 0 & 0 & 0 & 0 & 0 & 0 & 0 & 0 & 0 \\
\hline
	vanilla & ELU & 0.01 & 2593 & 0 & 0 & 0 & 0 & 0 & 0 & 0 & 0 & 0 & 0 & 0 & 0 & 0 & 0 \\
	vanilla & ELU & 0.1 & 874 & 1009 & 4166 & 5737 & 0 & 0 & 0 & 0 & 0 & 0 & 0 & 0 & 0 & 0 & 0 \\
	vanilla & ELU & 1 & 37 & 204 & 312 & 538 & 1507 & 1625 & 1642 & 1760 & 0 & 3265 & 0 & 4702 & 5331 & 0 & 0 \\
	vanilla & ELU & 5 & 0 & 0 & 0 & 1564 & 815 & 984 & 364 & 461 & 520 & 653 & 631 & 783 & 841 & 0 & 1201 \\
	vanilla & sigmoid & 0.01 & 0 & 0 & 0 & 0 & 0 & 0 & 0 & 0 & 0 & 0 & 0 & 0 & 0 & 0 & 0 \\
	vanilla & sigmoid & 0.1 & 2406 & 0 & 0 & 0 & 0 & 0 & 0 & 0 & 0 & 0 & 0 & 0 & 0 & 0 & 0 \\
	vanilla & sigmoid & 1 & 133 & 0 & 3626 & 4204 & 0 & 0 & 0 & 0 & 0 & 0 & 0 & 0 & 0 & 0 & 0 \\
	vanilla & sigmoid & 5 & 127 & 1466 & 1930 & 2907 & 2613 & 3115 & 3930 & 0 & 2925 & 0 & 4598 & 3429 & 4273 & 0 & 4658 \\
	vanilla & TANH & 0.01 & 0 & 0 & 0 & 0 & 0 & 0 & 0 & 0 & 0 & 0 & 0 & 0 & 0 & 0 & 0 \\
	vanilla & TANH & 0.1 & 0 & 0 & 0 & 0 & 0 & 0 & 0 & 0 & 0 & 0 & 0 & 0 & 0 & 0 & 0 \\
	vanilla & TANH & 1 & 88 & 0 & 2873 & 2164 & 3583 & 4798 & 5304 & 4920 & 0 & 6055 & 0 & 0 & 7786 & 9281 & 8905 \\
	vanilla & TANH & 5 & 0 & 0 & 0 & 2020 & 1660 & 1006 & 1106 & 1133 & 1476 & 1291 & 1590 & 0 & 1487 & 0 & 1505 \\
	vanilla & L-R & 0.01 & 0 & 0 & 0 & 0 & 0 & 0 & 0 & 0 & 0 & 0 & 0 & 0 & 0 & 0 & 0 \\
	vanilla & L-R & 0.1 & 356 & 795 & 0 & 0 & 0 & 0 & 0 & 0 & 0 & 0 & 0 & 0 & 0 & 0 & 0 \\
	vanilla & L-R & 1 & 1345 & 1863 & 1537 & 4281 & 0 & 0 & 0 & 0 & 0 & 0 & 0 & 0 & 0 & 0 & 0 \\
	vanilla & L-R & 5 & 5965 & 0 & 0 & 0 & 6479 & 6129 & 5057 & 3507 & 3334 & 4238 & 4865 & 3666 & 3624 & 3512 & 3958 \\
	vanilla & RELU & 0.01 & 0 & 0 & 0 & 0 & 0 & 0 & 0 & 0 & 0 & 0 & 0 & 0 & 0 & 0 & 0 \\
	vanilla & RELU & 0.1 & 180 & 0 & 0 & 0 & 0 & 0 & 0 & 0 & 0 & 0 & 0 & 0 & 0 & 0 & 0 \\
	vanilla & RELU & 1 & 33 & 0 & 0 & 5089 & 0 & 0 & 0 & 0 & 0 & 0 & 0 & 0 & 0 & 0 & 0 \\
	vanilla & RELU & 5 & 0 & 0 & 0 & 0 & 0 & 0 & 0 & 0 & 5520 & 4815 & 5128 & 5004 & 4059 & 4258 & 4693 \\
	vanilla & B-R & 0.01 & 0 & 0 & 0 & 0 & 0 & 0 & 0 & 0 & 0 & 0 & 0 & 0 & 0 & 0 & 0 \\
	vanilla & B-R & 0.1 & 0 & 0 & 0 & 0 & 0 & 0 & 0 & 0 & 0 & 0 & 0 & 0 & 0 & 0 & 0 \\
	vanilla & B-R & 1 & 0 & 0 & 0 & 0 & 0 & 0 & 0 & 0 & 0 & 0 & 0 & 0 & 0 & 0 & 0 \\
	vanilla & B-R & 5 & 0 & 0 & 0 & 0 & 0 & 0 & 0 & 0 & 0 & 6259 & 0 & 6805 & 5126 & 0 & 4837 \\
	vanilla & LELU & 0.01 & 0 & 0 & 0 & 0 & 0 & 0 & 0 & 0 & 0 & 0 & 0 & 0 & 0 & 0 & 0 \\
	vanilla & LELU & 0.1 & 0 & 1495 & 3237 & 5311 & 0 & 0 & 0 & 0 & 0 & 0 & 0 & 0 & 0 & 0 & 0 \\
	vanilla & LELU & 1 & 229 & 1585 & 1082 & 600 & 1746 & 1873 & 3197 & 3195 & 3526 & 4984 & 7293 & 6816 & 8948 & 0 & 0 \\
	vanilla & LELU & 5 & 0 & 0 & 0 & 0 & 859 & 1494 & 1109 & 1427 & 654 & 1040 & 855 & 1213 & 1355 & 1788 & 1513 \\
	vanilla & LLELU & 0.01 & 2227 & 0 & 0 & 0 & 0 & 0 & 0 & 0 & 0 & 0 & 0 & 0 & 0 & 0 & 0 \\
	vanilla & LLELU & 0.1 & 1628 & 2142 & 4005 & 5888 & 0 & 0 & 0 & 0 & 0 & 0 & 0 & 0 & 0 & 0 & 0 \\
	vanilla & LLELU & 1 & 0 & 444 & 437 & 901 & 1221 & 1852 & 2948 & 2806 & 0 & 4703 & 5392 & 0 & 7256 & 0 & 8778 \\
	vanilla & LLELU & 5 & 2570 & 0 & 0 & 0 & 0 & 0 & 0 & 520 & 516 & 633 & 789 & 1158 & 1004 & 0 & 1060 \\
	vanilla & L3ELU & 0.01 & 4982 & 0 & 0 & 0 & 0 & 0 & 0 & 0 & 0 & 0 & 0 & 0 & 0 & 0 & 0 \\
	vanilla & L3ELU & 0.1 & 504 & 1132 & 5459 & 7705 & 0 & 0 & 0 & 0 & 0 & 0 & 0 & 0 & 0 & 0 & 0 \\
	vanilla & L3ELU & 1 & 377 & 152 & 454 & 614 & 998 & 1590 & 2075 & 2110 & 2784 & 0 & 4823 & 0 & 8018 & 0 & 0 \\
	vanilla & L3ELU & 5 & 0 & 0 & 0 & 1329 & 1535 & 469 & 535 & 445 & 633 & 756 & 751 & 0 & 987 & 0 & 1097 \\
\hline
    \hline
  \end {tabular}
\end{tiny}
  \end {table}

  \begin {table}
  \label{table:3d}
\begin{tiny}
  \begin{tabular}{ ||c|c|c||c|c|c|c|c|c|c|c|c|c|c|c|c|c|c||}
    \hline
    \hline
	Method & Activation & Rate & 2 & 3 & 5 & 7 & 11 & 13 & 17 & 19 & 23 & 29 & 31 & 37 & 41 & 43 & 47 \\
\hline
    \hline
	adadelta & ELU & 0.01 & 0 & 0 & 0 & 0 & 0 & 0 & 0 & 0 & 0 & 0 & 0 & 0 & 0 & 0 & 0 \\
	adadelta & ELU & 0.1 & 5465 & 8775 & 0 & 0 & 0 & 0 & 0 & 0 & 0 & 0 & 0 & 0 & 0 & 0 & 0 \\
	adadelta & ELU & 1 & 1321 & 2232 & 2730 & 3050 & 3096 & 4159 & 3386 & 3623 & 4375 & 4417 & 4524 & 6020 & 5097 & 0 & 0 \\
	adadelta & ELU & 5 & 351 & 1405 & 1149 & 1254 & 1310 & 1354 & 1202 & 1206 & 1319 & 1327 & 0 & 1915 & 1410 & 0 & 0 \\
	adadelta & sigmoid & 0.01 & 0 & 0 & 0 & 0 & 0 & 0 & 0 & 0 & 0 & 0 & 0 & 0 & 0 & 0 & 0 \\
	adadelta & sigmoid & 0.1 & 0 & 0 & 0 & 0 & 0 & 0 & 0 & 0 & 0 & 0 & 0 & 0 & 0 & 0 & 0 \\
	adadelta & sigmoid & 1 & 1410 & 4821 & 5495 & 0 & 7772 & 0 & 0 & 0 & 0 & 0 & 0 & 0 & 0 & 0 & 0 \\
	adadelta & sigmoid & 5 & 402 & 1509 & 4508 & 2897 & 4386 & 4590 & 0 & 2770 & 4029 & 0 & 4105 & 5327 & 0 & 5148 & 5300 \\
	adadelta & TANH & 0.01 & 0 & 0 & 0 & 0 & 0 & 0 & 0 & 0 & 0 & 0 & 0 & 0 & 0 & 0 & 0 \\
	adadelta & TANH & 0.1 & 0 & 0 & 0 & 0 & 0 & 0 & 0 & 0 & 0 & 0 & 0 & 0 & 0 & 0 & 0 \\
	adadelta & TANH & 1 & 1724 & 0 & 4698 & 6133 & 6315 & 6280 & 6793 & 0 & 0 & 7440 & 7013 & 0 & 0 & 0 & 0 \\
	adadelta & TANH & 5 & 0 & 1331 & 4062 & 2083 & 1779 & 1745 & 1678 & 0 & 1533 & 1641 & 1962 & 2859 & 2004 & 0 & 0 \\
	adadelta & L-R & 0.01 & 0 & 0 & 0 & 0 & 0 & 0 & 0 & 0 & 0 & 0 & 0 & 0 & 0 & 0 & 0 \\
	adadelta & L-R & 0.1 & 4853 & 0 & 0 & 0 & 0 & 0 & 0 & 0 & 0 & 0 & 0 & 0 & 0 & 0 & 0 \\
	adadelta & L-R & 1 & 2145 & 4578 & 4154 & 4119 & 5478 & 4976 & 6074 & 5482 & 5484 & 5849 & 7570 & 7643 & 7430 & 0 & 7734 \\
	adadelta & L-R & 5 & 1089 & 1609 & 2506 & 2303 & 3083 & 2076 & 2465 & 1768 & 1894 & 1661 & 1780 & 1966 & 1928 & 1825 & 1919 \\
	adadelta & RELU & 0.01 & 0 & 0 & 0 & 0 & 0 & 0 & 0 & 0 & 0 & 0 & 0 & 0 & 0 & 0 & 0 \\
	adadelta & RELU & 0.1 & 0 & 0 & 0 & 0 & 0 & 0 & 0 & 0 & 0 & 0 & 0 & 0 & 0 & 0 & 0 \\
	adadelta & RELU & 1 & 0 & 2612 & 4003 & 0 & 5285 & 6572 & 6297 & 5820 & 0 & 6323 & 6372 & 0 & 7914 & 0 & 8298 \\
	adadelta & RELU & 5 & 0 & 0 & 3214 & 3295 & 2283 & 3032 & 2011 & 2114 & 1665 & 1663 & 1607 & 1909 & 1833 & 1867 & 1985 \\
	adadelta & B-R & 0.01 & 0 & 0 & 0 & 0 & 0 & 0 & 0 & 0 & 0 & 0 & 0 & 0 & 0 & 0 & 0 \\
	adadelta & B-R & 0.1 & 0 & 0 & 0 & 0 & 0 & 0 & 0 & 0 & 0 & 0 & 0 & 0 & 0 & 0 & 0 \\
	adadelta & B-R & 1 & 0 & 0 & 0 & 0 & 0 & 0 & 0 & 0 & 0 & 0 & 0 & 0 & 0 & 0 & 0 \\
	adadelta & B-R & 5 & 0 & 0 & 0 & 0 & 0 & 0 & 0 & 0 & 3339 & 2964 & 2630 & 0 & 0 & 0 & 0 \\
	adadelta & LELU & 0.01 & 0 & 0 & 0 & 0 & 0 & 0 & 0 & 0 & 0 & 0 & 0 & 0 & 0 & 0 & 0 \\
	adadelta & LELU & 0.1 & 0 & 8581 & 0 & 0 & 0 & 0 & 0 & 0 & 0 & 0 & 0 & 0 & 0 & 0 & 0 \\
	adadelta & LELU & 1 & 0 & 2121 & 3555 & 3841 & 3849 & 4342 & 4965 & 5117 & 0 & 0 & 5198 & 0 & 6622 & 0 & 8025 \\
	adadelta & LELU & 5 & 568 & 1590 & 1707 & 1540 & 2274 & 1950 & 1906 & 1770 & 1810 & 1689 & 1747 & 1918 & 1929 & 2163 & 1936 \\
	adadelta & LLELU & 0.01 & 0 & 0 & 0 & 0 & 0 & 0 & 0 & 0 & 0 & 0 & 0 & 0 & 0 & 0 & 0 \\
	adadelta & LLELU & 0.1 & 7122 & 8577 & 0 & 0 & 0 & 0 & 0 & 0 & 0 & 0 & 0 & 0 & 0 & 0 & 0 \\
	adadelta & LLELU & 1 & 1147 & 2384 & 3396 & 5109 & 5237 & 5767 & 4541 & 5102 & 0 & 5883 & 6640 & 0 & 7290 & 0 & 0 \\
	adadelta & LLELU & 5 & 860 & 820 & 1806 & 2288 & 1946 & 2399 & 1810 & 1842 & 0 & 1895 & 2189 & 2185 & 2077 & 0 & 2226 \\
	adadelta & L3ELU & 0.01 & 0 & 0 & 0 & 0 & 0 & 0 & 0 & 0 & 0 & 0 & 0 & 0 & 0 & 0 & 0 \\
	adadelta & L3ELU & 0.1 & 4103 & 8360 & 0 & 0 & 0 & 0 & 0 & 0 & 0 & 0 & 0 & 0 & 0 & 0 & 0 \\
	adadelta & L3ELU & 1 & 1312 & 2635 & 3065 & 3777 & 3797 & 0 & 4327 & 0 & 0 & 0 & 5257 & 7806 & 0 & 0 & 0 \\
	adadelta & L3ELU & 5 & 615 & 1327 & 1201 & 1374 & 1609 & 1551 & 1463 & 1540 & 0 & 1565 & 1584 & 2098 & 1899 & 0 & 1914 \\
\hline
    \hline
  \end {tabular}
\end{tiny}

  \caption {Average number of epochs needed to solve exactly the generalized XOR classification problem 
with p classes, using different optimization strategies, activation functions and learning rates. An epoch
contains $10 p^2$ training examples. For each activation function, 10 tests were conducted and the results were averaged over at least 5 successful trials. See figure 2 for a definition of the activation functions.  The tests were conducted systematically up to $p=61$, but presented only up to $p=47$ to fit in the page, but the trends do not change in the $47$ - $61$ region. We explored a few cases up to 191, but the calculations become slow.
A zero indicates a case where less than 5 trials out of 10 converged in less than 10000 epochs.}
  \end {table}
 
\end{addmargin}

\end {document}